# Sentiment analysis with adaptive multi-head attention in Transformer

Fanfei Meng*(corresponding), Chen-Ao Wang
fanfeimeng2023@u.northwester.edu, wangcha@tju.edu.cn
Northwestern University, 633 Clark Street, Evanston, IL, US
Tianjin University, Tianjin, China

## 1. Abstract

We propose a novel framework based on the attention mechanism to identify the sentiment of a movie review document. Previous efforts on deep neural networks with attention mechanisms focus on encoder and decoder with fixed numbers of multi-head attention. Therefore, we need a mechanism to stop the attention process automatically if no more useful information can be read from the memory.In this paper, we propose an adaptive multi-head attention architecture (AdaptAttn) which varies the number of attention heads based on length of sentences. AdaptAttn has a data preprocessing step where each document is classified into any one of the three bins small, medium or large based on length of the sentence. The document classified as small goes through two heads in each layer, the medium group passes four heads and the large group is processed by eight heads. We examine the merit of our model on the Stanford large movie review dataset. The experimental results show that the F1 score from our model is on par with the baseline model.

## 2. Introduction

The essence of the text sentiment analysis task is feature extraction of natural language sequence and feature-based classification task. Text feature extraction has always been the main research direction of Natural Language Processing (NLP). From Recurrent Neural Network (RNN) to AutoEncoder to Bidirectional Encoder Representations from Transformers (BERT), improvements have been made in the methods of feature extraction. The idea of this paper is similar to BERT: Encoder of Transformer is used as a feature extractor and then connected with a fully connected neural network for classification fitting.

The multi-head self-attention module is a key component in Transformer. Like most language models, the Transformer (which will be referred to as "vanilla Transformer" to distinguish it from other enhanced versions) are trained on sentences with a fixed length and the sentences shorter than the length will be padded with too many zeros. In this work, we dynamically changed the number of layers adapted to the length of the sentences to avoid some waste of computation and improve ability to process longer sentences. We analyze the dataset, and generate three ranges to use a different number of heads, which are 2, 4, and 8. The non-adaptive model will use 8 heads for the whole dataset. The results of our experiments show that both approaches of adaptive attention provide better accuracy compared to the benchmark. And according to the F1-score and accuracy, the adaptive models have comparable accuracy with the non-adaptive one but used much less training time.

## 3. Related Work

The sentiment analysis method based on deep learning firstly used a neural network to encode the text data, namely word. Then the encoding is input into the feature extraction network to extract the semantic and syntactic feature information of the text data. Finally, a classifier is used to obtain the emotional polarity of the text. [1] firstly used Convolution Neural Network (CNN) for sentiment analysis tasks, used only for the training of the word vector and standard convolutional neural network, and got better effect than traditional machine learning methods. [2] used a multi-channel convolutional neural network to improve the classification performance of emotional polarity of the model by adding feature information such as part of speech, position and emotion. [3] introduced a recursive neural network to conduct sentiment analysis on film review text data.

Attention Mechanism was first used in the field of images to enable neural networks to focus on certain information when processing data. Moreover, due to the high concurrent computing power, the attention mechanism is more and more widely used in natural language processing tasks. Self-attention is a type of attention mechanism where the model makes predictions for one part of a data sample using other parts of the observation about the same sample. There are various forms of attention / self-attention, Transformer [4] relies on the scaled dot-product attention. The multi-head self-attention module is a key component in Transformer. Rather than only computing the attention once, the multi-head mechanism splits the inputs into smaller chunks and then computes the scaled dot-product attention over each subspace in parallel. The Transformer model has an encoder-decoder architecture, as commonly used in many NMT models. Later decoder-only transformers were shown to achieve great performance in language modeling tasks, like in GPT[5] and BERT[6].

## 4. The Model

Transformer has demonstrated its abilities of sentence processing and recognition, including text classification, machine translation and text generation. It is a state-of-art architecture to take care of semantic analysis via parallel computing computations. Compared with traditional RNN structure, the multi-head attention mechanism can process the whole paragraph or multiple sentences in total to obtain their interrelationship. In Transformer, encoder positioner, multi-head attention mechanism and feedforward comprise a complete unit. The encoder position is for marking the word position in each sentence, which can effectively ensure the position of each word can be fully considered in sentence analysis. When it comes to the multi-head attention mechanism, there are three trainable matrices: k,q,v for dividing samples into multiple heads and analyzing their attention relationship separately under parallel computing. In the meanwhile, the word dependence will be preserved due to former positional encoding. Finally, the processed samples pass the feed-forward, which is composed of feed-forward networks to serve as gradient descent during the backpropagation.

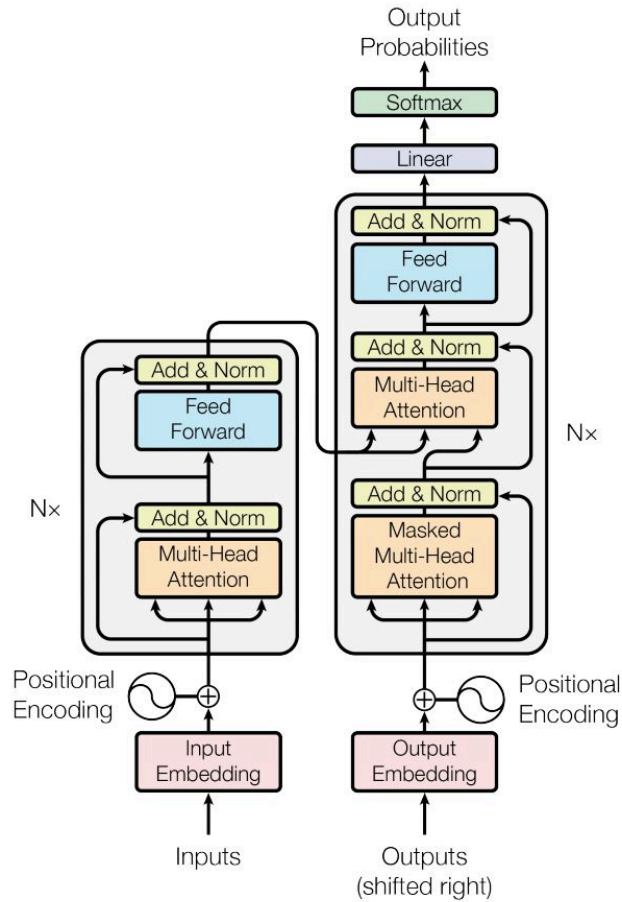

Figure 1: The visualized architecture of Transformer

The multi-head attention mechanism may cause overfitting or computing waste for short sentences, especially in the final training stage. Consequently, adaptive mechanisms have been widely applied in statistical language models for tackling various interesting issues. The core objective of the mechanism is to dynamically tune the architecture of the model to accommodate the context states, especially the issue that context states emphasize more flexible hyper parameters in the model.

Inspired by the two issues, we introduce a state-of-art adaptive Transformer, which is able to process sentences with different lengths using different numbers of heads. For text classification, the length of sentence is pivotal to the performance of the model, especially the Transformer. Transformer outperforms in short sentences; however, it is not fully adaptable to long sentences. On the other hand, if far more condensed attention is imposed on short sentences, overfitting will negatively impact the final recognition accuracy with training going through. To be specific, we initially extend the distribution of sentence length in our dataset and divide them into three groups : Small, Medium, Large with closed group size. After that, we apply sparse Attention to process Small Groups, Medium Attention to process Medium Groups, condense Attention to process Large Groups. The figure is shown below for Attention mechanism. In our model, we define condensed Attention contains 8 heads, medium Attention includes 4 heads and sparse Attention only has 2 heads.

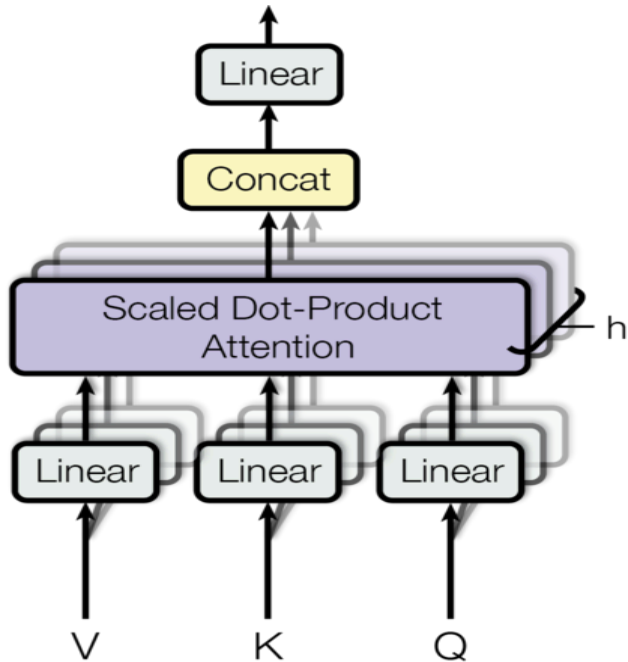

Figure 2: The visualized architecture of multi-head attention

After the Feed-forward network, the processing is to do mean-pooling on sequence length and impose a fully connected layer on embedding. Finally the softmax is for computing probabilities of each class. Pipeline is shown below:

**Dimension Flow:** (batchsize, sequence length, embedding size) - (batchsize, 1, embedding size) - (batchsize, 1, 2) - Softmax(batchsize, 1, 2) - (batchsize, P1, P2).

**Equation Flow:** Mean (x, dim=1) – Fully Connected Layer(x, dim=2) – Softmax(x, dim=2) – (P1 , P2).

## 5. Experimental Setting

We conducted experiments on Stanford's large movie review dataset which contains a set of 50,000 movie review documents classified either as positive or negative. We are using 25,000 documents as training data and the other 25,000 documents as test data. In our non-adaptive approach, all the 25,000 documents got through 8 attention heads.

We have two adaptive approaches with different percentages of documents in each bin. The notion behind is to study the impact of computation time and accuracy. In every adaptive approach there is a pre-processing step which we did manually by choosing two threshold values as lengths L1 and L2. All documents with length less than L1 are classified small, between L1 and L2 as medium and greater than L2 as large, the splits are as shown in Table 1. A document classified as small has 2 multi-head attention, medium has 4 and large has 8.

In AdaptAttn Approach1, documents with 0-75 words are small, 76-150 are medium and $>$150 iare large. With these ranges we had 8\% small documents, 32\% medium documents and 60\% large documents. In this majority of the examples train with multi-head attention 8.

In the second adaptive approach, we define a document with 0-110 words as small, 110-200 as medium and $>$ 200 as large. With these ranges, 15\% are small documents, 52\% of medium and the rest 33\% of large.

|  | L1 [word count] | L2 [word count] | \%Small [ 2 heads] | \%Medium [ 4 heads] | \%Large [ 8 heads] |
| --- | --- | --- | --- | --- | --- |
| Non-adaptive attention | - | - | - | - | 100 |
| AdaptAttn Approach1 | 75 | 150 | 8 | 32 | 60 |
| AdaptAttn Approach2 | 110 | 200 | 15 | 52 | 33 |

Table1: Details of the experimental approaches from Stanford's movie review dataset.

Main Results

The evaluation metric is F1-score and accuracy and Table 2 shows the impact of AdaptAttn approaches. The previously published benchmark on the Stanford Large Movie review dataset has an accuracy of 88.89 and this was achieved using "bag of words". The non-adaptive attention trained by us on the same dataset has accuracy of 91.80 which is more relevant for comparison to the adaptive model.

From our results, we can say that both approaches of adaptive attention provide better accuracy compared to the benchmark. Approach 1 has a ~6 unit increase compared to the non-adaptive attention while approach 2 has only ~1 unit increase. We think approach 2 accuracy is less because the majority of training examples had 4 heads. We also observe that training time increases with multi-attention heads which is as expected. The surprising result was that adaptive approach 1 performs better than non-adaptive which we would like to analyze further.

|  | F1-Score | Accuracy (%) | Training Time |
| --- | --- | --- | --- |
| Non-adaptive attention | 0.9107 | 91.80 | ~120 epochs in 30hrs |

| Approach 1 | 0.9714 | 97.22 | ~140 epochs in 30hrs |
| Approach 2 | 0.9236 | 92.12 | ~170 epochs in 30hrs |

Table 2: The results of our non-adaptive and adaptive approaches.

# 6. Conclusion and Future Work

In this paper, we proposed a framework to vary attention heads based on size of the document. The model first runs through the input to generate a memory, in the process of which it can synthesize the word sequence features. And then, the model pays adaptive attention on the memory to pick up important information to predict the final sentiment, by combining the features from different attentions non-linearly. We demonstrated the efficacy of our model on Stanford Large Movie Review datasets, and the results show that it can be in par with the state-of-the-art methods.Further, in sentiment classification a mechanism to get raw F1 scores bin wise would provide more insights into results. We would also like to expand adaptive attention to other tasks like machine translation.

# References

[1] Kim Y . Convolutional Neural Networks for Sentence Classification[J]. Eprint Arxiv, 2014.
[2] Chen, K.; Liang, B.; Ke, W.; Xu, B.; Zeng, G.C. Chinese Micro—Blog Sentiment Analysis Based on Multi-Channels Convolutional Neural Networks. J. Comput. Res. Dev. 2018, 55, 945–957.
[3] Richard Socher, Brody Huval, Christopher D. Manning, and Andrew Y. Ng. 2012. Semantic compositionality through recursive matrix-vector spaces. In Proceedings of the Conference on Empirical Methods in Natural Language Processing, pages 1201–1211.
[4] Vaswani A , Shazeer N , Parmar N , et al. Attention Is All You Need[J]. arXiv, 2017.
[5] Radford, A., Narasimhan, K., Salimans, T. and Sutskever, I., 2018. Improving language understanding by generative pre-training.
[6] Devlin, J., Chang, M.W., Lee, K. and Toutanova, K., 2018. Bert: Pre-training of deep bidirectional transformers for language understanding.
[7] Maas, A.L., Daly, R.E., Pham, P.T., Huang, D., Ng, A.Y. and Potts, C., 2011, June. Learning word vectors for sentiment analysis. In Proceedings of the 49th annual meeting of the association for computational linguistics: Human language technologies-volume 1 (pp. 142-150). Association for Computational Linguistics.
[8] Branden Ghena Fanfei Meng. Research on text recognition methods based on artificial intelligence and machine learning. preprint under review, 2023.
[9] Fanfei Meng and David Demeter. Sentiment analysis with adaptive multi-head attention in transformer, 2023.
[10] Manijeh Razeghi, Arash Dehzangi, Donghai Wu, Ryan McClintock, Yiyun Zhang, Quentin Durlin, Jiakai Li, and Fanfei Meng. Antimonite-based gap-engineered type-ii superlattice materials grown by mbe

and mocvd for the third generation of infrared imagers. In Infrared Technology and Applications XLV, volume 11002, pages 108–125. SPIE, 2019.

[11] Fanfei Meng, Lele Zhang, and Yu Chen. Fedemb: An efficient vertical and hybrid federated learning algorithm using partial network embedding.

[12] Fanfei Meng, Lele Zhang, and Yu Chen. Sample-based dynamic hierarchical trans-former with layer and head flexibility via contextual bandit.

[13] Fanfei Meng and Chen-Ao Wang. Adynamic interactive learning interface for computer science education: Program-ming decomposition tool.

[14] Chang Ling, Chonglei Zhang, Mingqun Wang, Fanfei Meng, Luping Du, and Xiaocong Yuan, "Fast structured illumination microscopy via deep learning," Photon. Res. 8, 1350-1359 (2020)

[15] Meng, Fanfei, Lalita Jagadeesan, and Marina Thottan. "Model-based reinforcement learning for service mesh fault resiliency in a web application-level." *arXiv preprint arXiv:2110.13621* (2021).

[16] Chen, Jin-Jin, et al. "A dataset of diversity and distribution of rodents and shrews in China." *Scientific Data* 9.1 (2022): 304

[17] Meng, F., Zhang, L., Wang, Y., & Zhao, Y. (2023). Joint detection algorithm for multiple cognitive users in spectrum sensing. *Authorea Preprints*.

[18]Meng, F., & Wang, Y. (2023). Transformers: Statistical interpretation, architectures and applications. *Authorea Preprints*.

[19]Fanfei Meng, Chen-Ao Wang, Alexander Brown. Evolution and Efficiency in Neural Architecture Search: Bridging the Gap Between Expert Design and Automated Optimization. *TechRxiv*. February 14, 2024.